\pdfoutput=1

\documentclass[11pt]{article}

\usepackage{acl}

\usepackage{times}
\usepackage{latexsym}

\usepackage[T1]{fontenc}

\usepackage[utf8]{inputenc}

\usepackage{microtype}

\usepackage{hyperref}
\usepackage{url}
\usepackage{pifont}
\usepackage{url}
\usepackage[utf8]{inputenc} 
\usepackage[T1]{fontenc}    
\usepackage{url}            
\usepackage{booktabs}       
\usepackage{amsfonts}       
\usepackage{nicefrac}       
\usepackage{microtype}      
\usepackage{times}
\usepackage{xcolor}
\usepackage{color}
\usepackage{soul}
\usepackage[utf8]{inputenc}
\usepackage{graphicx} 
\usepackage{wrapfig}
\usepackage{amssymb}
\usepackage{amsmath}
\usepackage{multirow}
\usepackage{float}
\usepackage{subcaption}
\usepackage{arydshln}

\usepackage{times}
\usepackage{epsfig}
\usepackage{graphicx}
\usepackage{amsmath}
\usepackage{amssymb}
\usepackage{textcomp}

\usepackage[ruled,vlined,linesnumbered]{algorithm2e}

\DeclareMathOperator*{\argmax}{arg\,max}

\title{$C^3$: Compositional Counterfactual Contrastive Learning for Video-grounded Dialogues}

\author{Hung Le \\
  Salesforce Research Asia \\
  \texttt{hungle@salesforce.com} \\\And
  Nancy F. Chen \\
  A*STAR, Institute for Infocomm Research \\
  \texttt{nfychen@i2r.a-star.edu.sg} \\\AND
  Steven C.H. Hoi \\
  Salesforce Research Asia \\
  \texttt{shoi@salesforce.com} \\}

\begin{document}
\maketitle

\begin{abstract}
Video-grounded dialogue systems aim to integrate video understanding and dialogue understanding to generate responses that are relevant to both the dialogue and video context. Most existing approaches employ deep learning models and have achieved remarkable performance, given the relatively small datasets available. However, the results are partially accomplished by exploiting biases in the datasets rather than developing multimodal reasoning, resulting in limited generalization. In this paper, we propose a novel approach of Compositional Counterfactual Contrastive Learning ($C^3$) to develop contrastive training between factual and counterfactual samples in video-grounded dialogues. Specifically, we design factual/counterfactual samples based on the temporal steps in videos and tokens in dialogues and propose contrastive loss functions that exploit object-level or action-level variance. Different from prior approaches, we focus on contrastive hidden state representations among compositional output tokens to optimize the representation space in a generation setting. We achieved promising performance gains on the Audio-Visual Scene-Aware Dialogues (AVSD) benchmark and showed the benefits of our approach in grounding video and dialogue context. 
\end{abstract}
\section{Introduction} 
\begin{figure}[htbp]
	\centering
	\resizebox{1.0\columnwidth}{!} {
	\includegraphics{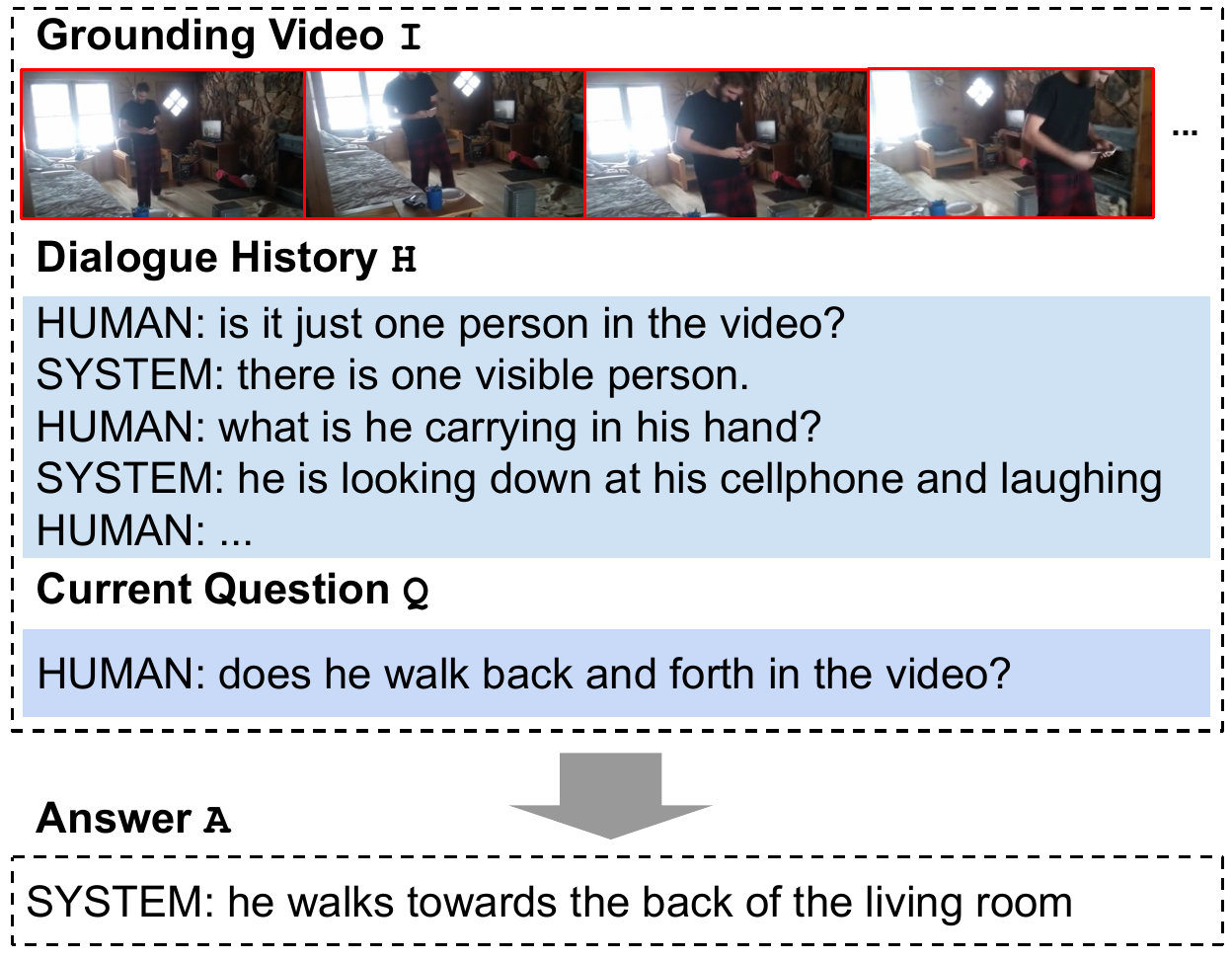}
	}
	\caption{
    An example of video-grounded dialogue
	}
	\label{fig:dial_example}
	\vspace{-0.1in}
\end{figure}

Visual dialogue research \citep{das2017visual, seo2017visual, de2017guesswhat, visdial_eval, alamri2019audio} aims to develop intelligent systems that can reason and answer questions about visual content in a multi-turn setting.
Compared to traditional visual question answering (VQA) \citep{antol2015vqa, gao2015you, malinowski2014multi, zhu2016visual7w}, visual dialogues bridge the gap between research and practical applications by allowing turn-based human-machine interactions.
Recently, many deep learning approaches have been proposed to develop visual dialogue systems and achieved remarkable performance \cite{schwartz2019factor, hori2019avsd, le-etal-2019-multimodal, 9376902}.
However, as these methods are heavily trained on relatively small datasets \citep{das2017visual, alamri2019audio}, 
they are subject to inherent bias from the datasets and limited generalization into real-world applications \cite{zhang2016yin, goyal2017making}. 
While training on large-scale data can alleviate this problem, visual dialogues are expensive to procure and require manual annotations.
This challenge becomes more obvious in highly complex visual dialogue tasks such as video-grounded dialogues \cite{alamri2019audio, le2021dvd} (Figure \ref{fig:dial_example}). 

In recent years, we have seen increasing research efforts in contrastive learning to improve deep learning performance \citep{wu2018unsupervised, henaff2020data, chen2020simple, he2020momentum}. 
The common strategy of these methods is an objective function that pulls together representations of an anchor and ``positive'' samples while pushing the representations of the anchor from ``negative'' samples.
These methods are specifically beneficial in self-supervised image representation learning.
Specifically, these methods often do not require additional annotations by augmenting data of existing samples to create ``positive'' and ``negative'' samples.
We are motivated by this line of research to improve visual dialogue systems and propose a framework of \textbf{\underline{C}}ompositional \textbf{\underline{C}}ounterfactual \textbf{\underline{C}}ontrastive Learning ($C^3$).
$C^3$ includes loss functions that exploit contrastive training samples of factual and counterfactual data that are augmented to be object-variant or action-variant. 

Compared to traditional deep learning tasks, a major challenge of applying contrastive learning \citep{wu2018unsupervised, henaff2020data, chen2020simple, he2020momentum} in video-grounded dialogues lies in the complexity of the task.
Specifically, in a discrimination task of image classification, given an image, positive samples are created based on non-adversarial transformations on this image e.g. by cropping inessential parts without changing the labels, and negative samples are randomly sampled from other image instances.
However, such transformations are not straightforward to apply on visual dialogues, each of which consists of a video of spatio-temporal dimensions, a dialogue of multiple turns, and an output label in the form of natural language at the sentence level. 
In visual dialogues, the random sampling method, in which negative samples are created by swapping the input video and/or dialogue context with random components from other training samples, becomes too naive.
In domains with high data variance like dialogues or videos, a system can easily discriminate between such positive and negative instances derived using previous approaches. 

\begin{figure*}[htbp]
	\centering
	\resizebox{1.0\textwidth}{!} {
	\includegraphics{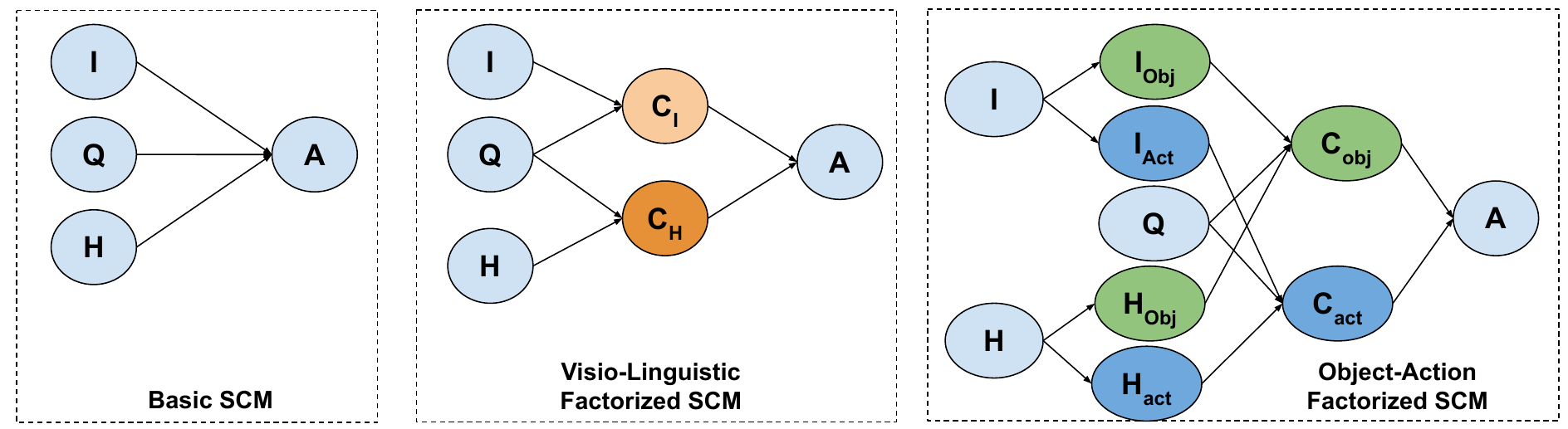}
	}
	\caption{
	\textbf{SCMs of video-grounded dialogues:}
	Left: Basic SCM without factorization.
	Middle: SCM factorized by visual and textual context. 
	Right: SCM factorized by object and action-level information. 
	I: video input, Q: question input, H: dialogue history, C: contextualized information, and A: target response. For simplicity, we do not demonstrate independent noise variables $U$ and the subscript $t$.
	}
	\label{fig:scm_models}
\end{figure*}

To mitigate the limitations of conventional contrastive learning in video-grounded dialogues, we propose a principled approach to generate and control negative and positive pairs by incorporating compositionality and causality (an overview of our approach can be seen in Figure \ref{fig:scm_models} and \ref{fig:cf_augmentation}).
Specifically, we develop a structural causal model for visual dialogues by decomposing model components by object and action-based aspects. 
We then create hard negative samples of grounding videos by masking temporal steps that are relevant to actions mentioned in target output responses. 
Hard negative dialogue samples are created by masking tokens that are referenced to the entity mentioned in target output responses. 
Positive samples of videos and dialogues are developed similarly by masking irrelevant temporal steps or tokens for them to remain factual. 
Finally, based on an object or action-based variance between factual and counterfactual pairs, we only select specific hidden state representations of the target dialogue response sequence, to apply contrastive loss functions.
Compared to existing approaches, our method has better control of data contrast at the granularity of object and action variance.
We conducted experiments with comprehensive ablation analysis using the Audio-Visual Scene-Aware Dialogues (AVSD) benchmark \cite{alamri2019audio}
and showed that our method can achieve promising performance gains.

\section{Related Work}
\textbf{Counterfactual Reasoning.}
Related to our work is the research of counterfactual reasoning.
One line of research focuses on generating plausible counterfactual data to facilitate model training or evaluation. 
\cite{zmigrod-etal-2019-counterfactual, garg2019counterfactual, NEURIPS2020_92650b2e} introduced data augmentation methods that convert gender-inflected sentences or remove identity-based tokens from sentences.
The augmented data is used to study model stereotyping and improve fairness in model outputs. 
\cite{Kaushik2020Learning} crowd-sourced human annotations to minimally revise documents such that their sentiment labels are flipped.  
\cite{zeng-etal-2020-counterfactual, wang2020robustness, madaan2020generate} introduced data augmentation to improve model robustness in entity recognition and text classification tasks. 

More related to our work are counterfactual augmentation methods in generative tasks. 
\cite{qin-etal-2019-counterfactual} introduced a new benchmark for counterfactual story rewriting.
\cite{li2021coco} explored augmented counterfactual dialogue goals to evaluate dialogue state tracking models.
\cite{Baradel2020CoPhy} proposed a synthetic 3D environment for learning the physical dynamics of objects in counterfactual scenarios. 
Different from prior tasks, in the task of video-grounded dialogue, a target response is not easy to be flipped/negated, and hence, supervised learning is not straightforward. 
We propose to automatically develop counterfactual and factual samples and improve representation learning via unsupervised learning. 

\textbf{Contrastive Learning.} 
Our work is related to the research of contrastive learning in deep learning models. 
The research is particularly popular in self-supervised learning of image representations \cite{wu2018unsupervised, hjelm2018learning, henaff2020data, chen2020simple, he2020momentum, NEURIPS2020_d89a66c7}.
These methods do not require additional annotations but aim to improve representations through loss functions.
The loss functions are often inspired by noise contrastive estimation (NCE) \cite{pmlr-v9-gutmann10a} and applied in lower-dimensional representation space. 
In the language domain, similar loss functions have been introduced to improve word embeddings \cite{NIPS2013_db2b4182} and sentence embeddings \cite{logeswaran2018an}.
More related to our work is \cite{huang-etal-2018-large, liu2015contrastive, yang-etal-2019-reducing, lee2021contrastive}, introducing positive and negative pairs of sentences for contrastive learning in generative tasks such as language modelling, word alignment, and machine translation. 
In the multimodal research domains, our work is related to contrastive learning methods introduced by \cite{zhang2020counterfactual, gokhale-etal-2020-mutant, liang2020learning, gupta2020contrastive}. 
Specifically, our work complements \cite{zhang2020counterfactual} by incorporating causality into contrastive learning. However, we focus on a very different task of video-grounded dialogues that involves turn-based question-answering. 
The task requires multimodal reasoning performed on both dialogue context and video context. 
Moreover, we improve models by tightly controlling data variance by adopting compositionality and our loss functions optimize hidden state representations of decoding tokens by their object or action-based semantics.  

\section{Method}
\label{sec:method}
\subsection{Problem Definition}
In a video-grounded dialogue task \cite{alamri2019audio, le2021dvd}, the inputs consist of a dialogue $\mathcal{D}$ and the visual input of a video $\mathcal{I}$.
Each dialogue contains a sequence of dialogue turns, each of which is a pair of question $\mathcal{Q}$ and answer $\mathcal{A}$.
At each dialogue turn $t$, we denote the dialogue context $\mathcal{H}_t$ as all previous dialogue turns $\mathcal{H}_t=\{(\mathcal{Q}_i, \mathcal{A}_i)\}|_{i=1}^{i=t-1}$.
The output is the answer $\hat{\mathcal{A}}_t$ to answer the question of the current turn $\mathcal{Q}_t$. 
The objective of the task is the generation objective that output answers of the current turn:
\begin{align}
    \hat{\mathcal{A}}_t = \argmax_{\mathcal{A}_t} \displaystyle P(\mathcal{A}_t|\mathcal{I}, \mathcal{H}_t, \mathcal{Q}_t; \theta) \label{eq:obj_f}
\end{align}
\subsection{Structural Causal Model} 
\label{subsec:scm}
We first cast a visual dialogue model as a structural causal model (SCM) \cite{pearl2009causal} to explore the potential factors that affect the generation of target dialogue responses in a dialogue system.
By definition, an SCM consists of random variables $V=\{V_1,...,V_N\}$ and corresponding independent noise variables $U=\{U_1,...,U_N\}$.
We assume an SCM of a directed acyclic graph (DAG) structure. 
In this structure, causal functions are defined as $F=\{f_1,...,f_N\}$ such that $V_i=f_i(P_i, U_i)$ where $P_i=\{V_p\} \subset V$ are the parent nodes of $V_i$ in the DAG.   
Using this definition of SCM, we develop three SCM structures for a video-grounded dialogue system in Figure \ref{fig:scm_models}. 

\begin{figure*}[htbp]
	\centering
	\resizebox{1.0\textwidth}{!} {
	\includegraphics{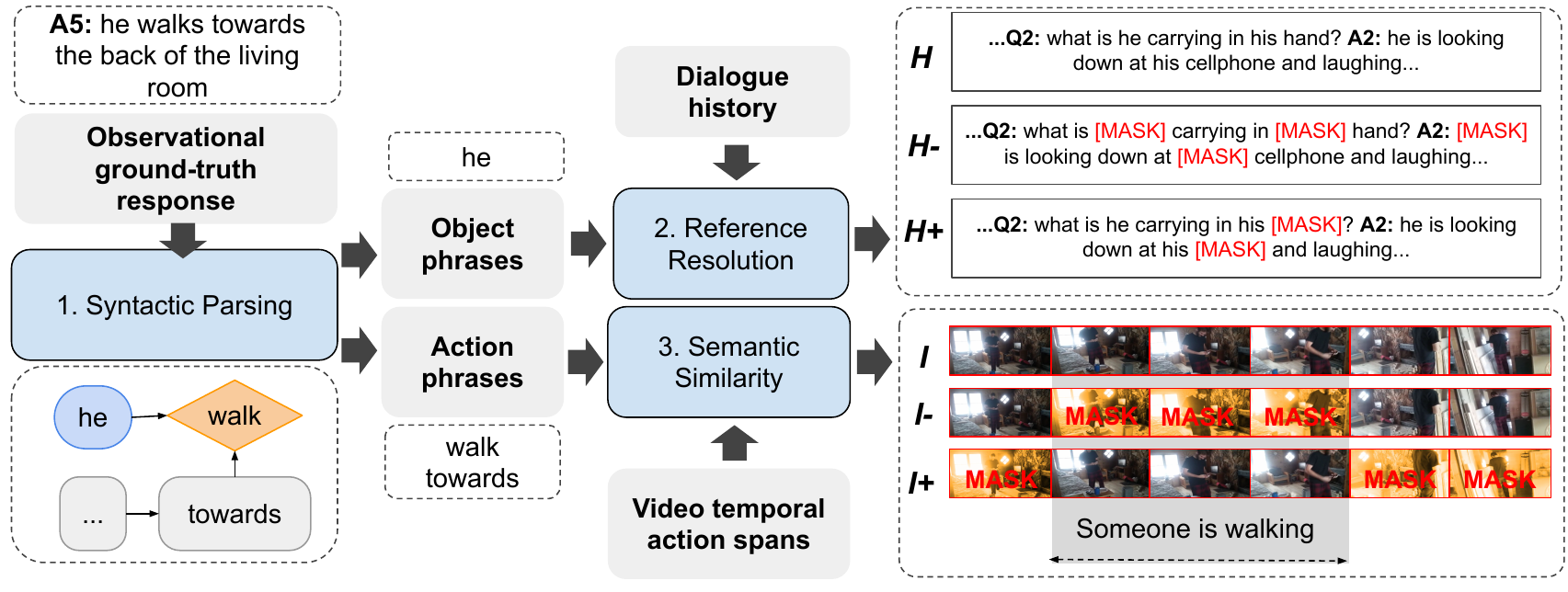}
	}
	\caption{
	\textbf{Counterfactual generation:}
	An overview of our factual and counterfactual dialogue/video generation.
	}
	\label{fig:cf_augmentation}
\end{figure*}

The \emph{Basic SCM} is directly derived from the objective function (\ref{eq:obj_f}).
The \emph{VL-SCM} adopts a question-aware reasoning process that partitions visual and language reasoning based on question information as the common cause.
A limitation of VL-SCM is that it does not account for the interactions of components such as object and action abstracts that are embedded in visual context $C_I$ and linguistic context $C_H$.
This drawback becomes more significant in scenarios in which question information is highly dependent on prior turns in the dialogue history. 
Specifically, in questions that involve references, including object references (``does \emph{she} interact with \emph{the woman in red}?'') and action references (``what does the boy do after \emph{that}?''), VL-SCM is not optimal to integrate dialogue and video context to solve component references such as ``she'' and ``that''.
To address this drawback, we propose an \emph{OA-SCM} that is factorized by object-action contextual information (Figure \ref{fig:scm_models}, right). 
The causal functions $f^{obj}_H$ and $f^{act}_H$ can be a simple text parser that map tokens into object-based tokens or action-based tokens s.t. $\mathcal{H}_{obj}=f^{obj}_H(\mathcal{H})$ and $\mathcal{H}_{act}=f^{act}_H(\mathcal{H})$.
Similarly, $f^{obj}_I$ and $f^{act}_I$ are causal functions that map bounding boxes or temporal steps into object-based or action-based contents. 
In Section \ref{subsec:cf_augmentation}, we show that OA-SCM structure provides a framework to develop \emph{partially counterfactual} training samples. 
\subsection{Counterfactual Augmentation}
\label{subsec:cf_augmentation}
An overview of our augmentation process can be seen in Figure \ref{fig:cf_augmentation}.

\textbf{Decomposing observational target response.}
First, at each dialogue turn $t$, the ground-truth dialogue response $\mathcal{A}_t$ are passed to a syntactic parser such as the Stanford parser system \footnote{\url{https://nlp.stanford.edu/software/lex-parser.shtml}}.
The output includes grammatical components, such as subjects, verbs, and modifiers, in the form of a dependency tree. 
We prune the dependency tree to remove inessential parts and extract a set of object phrases $\mathcal{A}_{t,obj}$, and action-based phrases $\mathcal{A}_{t,act}$.

\textbf{Generating counterfactual dialogue.}
Based on $\mathcal{A}_{t,obj}$, we apply a pretrained reference resolution model e.g. \cite{clark-manning-2016-deep}, to the dialogue context $\mathcal{H}_t$ to identify any references from past dialogue turns to any objects in $\mathcal{A}_{t,obj}$.
For instance, in Figure \ref{fig:scm_models}, the object ``he'' identified in $\mathcal{A}_t$ are mapped to different token positions in prior dialogue turns, e.g. ``his'' in the text span ``his hand'' in the second question turn.  
All referenced tokens in dialogue context $\mathcal{H}_t$ are replaced by a \texttt{MASK} vector and the resulting dialogue context is denoted as counterfactual sample $\mathcal{H}^{-}_t$.
We also used the pretrained reference resolution model to select any object tokens in $\mathcal{H}_t$ that are not mapped to $\mathcal{A}_{t,obj}$.
These objects are considered irrelevant to $\mathcal{A}_t$ and they are replaced by the \texttt{MASK} vector from $\mathcal{H}_t$ and the resulting dialogue is denoted as a factual sample $\mathcal{H}^{+}_t$.

\textbf{Generating counterfactual video.}
To create a counterfactual video sample, we first identify the temporal steps from the video that are semantically relevant to action phrases in $\mathcal{A}_{t,act}$.
We obtain the annotation of temporal action spans from video, which can be retrieved from a pretrained temporal localization model \cite{shou2016temporal} or is readily available in existing video benchmarks \cite{sigurdsson2016hollywood}. 
The action span annotations consist of a set of action labels $Y_{i,act}$, each of which is mapped to a start and end time $(t^{s}_i, t^{e}_i)$.
Temporal segments that are deemed necessary to generate $\mathcal{A}_t$ is the union of all time spans from the set $S=\{(t^{s}_j, t^{e}_j)\}$ for all $Y_{j,act}$ that is semantically similar to $\mathcal{A}_{t,act}$. 
To identify similar pairs, we adopted cosine similarity scores between pretrained Glove embedding vectors of $Y_{j,act}$ and  $\mathcal{A}_{t,act}$.
During video feature encoding, any features of temporal steps sampled within $S$ are replaced with a \texttt{MASK} vector, and resulting video features are noted as encoded features of counterfactual video $\mathcal{I}^{-}$.
Factual video $\mathcal{I}^{+}$ are created similarly but for video parts irrelevant to $\mathcal{A}_t$, that is $I\setminus S$.

By the definition of OA-SCM from Section \ref{subsec:scm}, we can denote $\mathcal{H}^{-}_t = \mathcal{H}^{-}_{t,obj} + \mathcal{H}_{t,act}$ and $\mathcal{H}^{+}_t = \mathcal{H}^{+}_{t,obj} + \mathcal{H}_{t,act}$; and
$\mathcal{I}^{-} = \mathcal{I}_{obj} + \mathcal{I}^{-}_{act}$ and $\mathcal{I}^{+} = \mathcal{I}_{obj} + \mathcal{I}^{+}_{act}$.
Note that we follow \cite{NEURIPS2018_496e05e1} and assume object information such as object appearance and shape are typically embedded in any video frame. 
In this case, $\mathcal{I}_{obj}$ is unchanged and can be obtained from either $I\setminus S$ or $S$. 
In Section \ref{subsec:contrast_learning}, we show that these partially counterfactual formulations enable a compositional contrastive learning approach.
\subsection{Contrastive Learning} 
\label{subsec:contrast_learning}
In this section, we introduce a contrastive learning method that exploits the compositional hidden states between factual and counterfactual samples. 
We extend the objective function (\ref{eq:obj_f}) to express the auto-regressive decoding process: 
\begin{align*}
    \hat{\mathcal{A}}_t &= \argmax_{\mathcal{A}_t} \displaystyle P(\mathcal{A}_t|\mathcal{I}, \mathcal{H}_t, \mathcal{Q}_t; \theta) \\
    &= \argmax_{\mathcal{A}_t} 
    \prod\limits_{m=1}^{L_\mathcal{A}}
    \displaystyle P_m(w_m| \mathcal{A}_{t,<m},\mathcal{I}, \mathcal{H}_t, \mathcal{Q}_t; \theta) 
\end{align*}
Each target response $\mathcal{A}$ is represented as a sequence of token or word indices $\{w_m\}|_{m=1}^{m=L} \in |\mathbb{V}|$, where $L$ is the sequence length and $\mathbb{V}$ is the vocabulary set. The conditional probability $P_m$ is defined as: 
\begin{align}
    \displaystyle P_m &= \mathrm{softmax} (W k_m + b) \in \mathbb{R}^{|\mathbb{V}|}\\
    k_m &= \theta_\mathrm{decode} (w_{m-1}, \theta_\mathrm{encode}(\mathcal{I}, \mathcal{H}_t, \mathcal{Q}_t)) 
    \label{eq:km}
\end{align}
where $k_m$ is the hidden state at decoding position $m$ and $d$ is the embedding dimension of the hidden state. 
In this generative setting, we then explain 2 different ways of contrastive learning:

\textbf{Sentence-level contrast}.
This approach learns the representations of the hidden states by contrasting a linear transformation of an aggregated vector of hidden states following an NCE framework: 
\begin{align}
    \mathcal{L}^\mathrm{sent}_\mathrm{nce} &= -\log \frac{e^{\mathrm{sim}(z, z^{+})}}
    {e^{\mathrm{sim}(z, z^{+})} + e^{\mathrm{sim}(z, z^{-})}} 
    \label{eq:sent_nce}
\end{align}
where $\mathrm{sim}(,)$ is the cosine similarity score and $z$ is the output of an aggregation function $\mathrm{Agg}$: $z = \mathrm{Agg}(U)$ where $U \in \mathbb{R}^{d_\mathrm{nce} \times L_\mathcal{A}}$ and $u_m = \mathrm{MLP_\mathrm{nce}}(k_m) \in \mathbb{R}^{d_\mathrm{nce}}$.
$z^+$ and $z^-$ are obtained similarly by passing $k^+_m$ and $k^-_m$ to the same MLP and aggregation function. 
$k^+_m$ and $k^-_m$  are obtained by passing factual and counterfactual video pairs into (\ref{eq:km}): 
$k^+_m = \theta_\mathrm{decode} (w_{m-1}, \theta_\mathrm{encode}(\mathcal{I}^+, \mathcal{H}_t, \mathcal{Q}_t))$
and
$k^-_m = \theta_\mathrm{decode} (w_{m-1}, \theta_\mathrm{encode}(\mathcal{I}^-, \mathcal{H}_t, \mathcal{Q}_t))$.
In cases of augmentation with factual and counterfactual dialogues, we obtain $k^+_m$ and $k^-_m$ by replacing $\mathcal{H}$ with $\mathcal{H}^+$ and $\mathcal{H}^-$ in (\ref{eq:km}). $\mathrm{Agg}$ is an aggregation function that collapses hidden states into a single vector, e.g. average pooling \cite{lee2021contrastive, zhang2020counterfactual}. We follow \cite{NEURIPS2020_d89a66c7} to normalize $z, z^+, z^-$ to lie on the unit hypersphere. 
To reflect this contrastive learning approach against the VL-SCM, we can assume $\mathcal{C} \cong K$ and (\ref{eq:sent_nce}) essentially exploits the contrast between $\mathcal{C}^+$ and $\mathcal{C}^-$.

\textbf{Compositional contrast}. 
We note that the above approach does not consider compositionality in the target output response $\mathcal{A}$.
Since we are using the same observational output $w_{m-1}$ to obtain $k_m$, $k^-_m$, and $k^-_m$, 
we can remove the $\mathrm{Agg}$ function and apply a token-level pairwise contrastive loss between pairs of $(z_m=u_m, z^+_m=u^+_m)$ and ($z_m=u_m$, $z^-_m=u^-_m)$.
In this strategy, we formulate a loss function for action variance between $\mathcal{I}^{+}$ and $\mathcal{I}^-$, and one for object variance between $\mathcal{H}^+$ and $\mathcal{H}^-$:
\begin{align}
    \mathcal{L}^\mathrm{act}_\mathrm{nce} &= - \frac{1}{|D_{act}|} \sum_{i \in D_{act}} \log \frac{e^{\mathrm{sim}(z_i, z^{+}_i)}}
    {e^{\mathrm{sim}(z_i, z^{+}_i)} + e^{\mathrm{sim}(z_i, z^{-}_i)}} \nonumber
    \\
    D_{act}&=\{\mathrm{idx}(w_i):w_{i-1} \in \mathcal{A}_{t,act}\}
    \label{eq:l_act}\\
    \mathcal{L}^\mathrm{obj}_\mathrm{nce} &= - \frac{1}{|D_{obj}|} \sum_{j \in D_{obj}} \log \frac{e^{\mathrm{sim}(z_j, z^{+}_j)}}
    {e^{\mathrm{sim}(z_j, z^{+}_j)} + e^{\mathrm{sim}(z_j, z^{-}_j)}}  \nonumber
    \\
    D_{obj}&=\{\mathrm{idx}(w_j):w_{j-1} \in \mathcal{A}_{t,obj}\}
    \label{eq:l_obj}
\end{align}
where $idx(w_m)$ returns the index of $w_m$ in $\mathcal{A}_t$. 
Note that in (\ref{eq:l_act}) and (\ref{eq:l_obj}), we adopt a hypothetical strategy by obtaining hidden states given \emph{input} tokens are either in $\mathcal{A}_{t,act}$ or $\mathcal{A}_{t,obj}$. 
An alternative approach is to consider hidden states that are expected to produce \emph{prospective} tokens $w_m \in \mathcal{A}_{t,act}$/$\mathcal{A}_{t,obj}$, i.e. $D'_{act}=\{\mathrm{index}(w_i):w_{i} \in \mathcal{A}_{t,act}\}$ and $D'_{obj}=\{\mathrm{index}(w_j):w_{j} \in \mathcal{A}_{t,obj}\}$.
We conducted experiments with both strategies and explained our findings in the next section.
Note that we can connect the compositional contrastive learning approach against the OA-SCM (Section \ref{subsec:scm}) by denoting $\mathcal{C}_{act} \cong \{k_i\} \forall i \in D_{act}$ and $\mathcal{C}_{obj} \cong \{k_j\} \forall j \in D_{obj}$.
Therefore, (\ref{eq:l_act}) essentially exploits the contrast between $\mathcal{C}^+_{act}$ and $\mathcal{C}^-_{act}$, and (\ref{eq:l_obj}) for the contrast between $\mathcal{C}^+_{obj}$ and $\mathcal{C}^-_{obj}$.

\begin{table*}[htbp]
\centering
\begin{tabular}{lccccccc}
\hline
            & \textbf{Train}  
            & \textbf{Train${^\mathrm{video}_\mathrm{aug}}$} & \textbf{Train${^\mathrm{dial}_\mathrm{aug}}$} & 
            \textbf{Val}    
            & \textbf{Val${^\mathrm{video}_\mathrm{aug}}$} & \textbf{Val${^\mathrm{dial}_\mathrm{aug}}$} 
            & \textbf{Test}  \\
            \hline
\textbf{\#Dialogs}  & 7,659  & 7,145                             & 6,411                             & 1,787  & 1,709                           & 1,557                          & 1,710 \\
\textbf{\#$(\mathcal{I},\mathcal{H}_t,\mathcal{Q}_t,\mathcal{A}_t)$} & 76,590 & 28,163                            & 18,397                            & 17,870 & 7,383                           & 4,912                          & 6,745\\
\hline
\end{tabular}
\caption{
Summary of the AVSD benchmark with augmented counterfactual video/dialogue data}
\label{tab:aug_data}
\end{table*}

\section{Experiments}
\label{sec:exp}
\textbf{Dataset and Experimental Setup}. We used the Audio-Visual Sene-Aware Dialogue (AVSD) dataset \cite{alamri2019audiovisual} to benchmark video-grounded dialogue systems. 
The dataset contains 10-turn dialogues, each of which is grounded on one video from the Charades dataset \citep{sigurdsson2016hollywood}.
We used the standard visual features I3D \cite{carreira2017quo} to represent the video input. 
Note that compared to \cite{alamri2019audiovisual}, we followed the setting of AVSD in the $7^{th}$ Dialogue System Technology Challenge (DSTC7) \citep{yoshino2019dialog}, which requires generating a response rather than selecting from a candidate set. 
We also did not use video caption as an input as the caption is typically not easy to obtain in applications.
A summary of the dataset can be seen in Table \ref{tab:aug_data}.

All model parameters, except the visual feature extractor of a pretrained I3D model, are initialized with uniform distribution \citep{glorot2010understanding}. 
Our approach can be applied to different model architectures, as long as the hidden states of individual decoding tokens are available for contrastive learning. 
We used MTN \cite{le-etal-2019-multimodal}, which is a Transformer adaptation of the traditional RNN-based dialogue systems, as our base model.
Finally, we evaluated models with objective metrics, including BLEU \citep{papineni2002bleu}, METEOR \citep{banerjee2005meteor}, ROUGE-L \citep{lin2004rouge}, and CIDEr \citep{vedantam2015cider}.
These metrics are found to correlate well with human judgment \citep{alamri2019audiovisual}.

\textbf{Creating Counterfactual Data}.
We created counterfactual data for the training split and validation split of the AVSD benchmark.
Specifically, from the original data, we identified invalid samples that are not sufficient for factual and counterfactual transformations. 
Examples of invalid samples are ones with ambiguous actions in target responses (e.g. ``I am not sure what he is doing''), or ones without object references to prior turns (e.g. ``there is only a single person in the video''). 
These samples are discarded and the remaining data is processed as described in Section \ref{subsec:cf_augmentation}.
The overall statistics of augmented train and validation splits can be seen in Table \ref{tab:aug_data}.
Note that the number of samples with augmented videos and dialogues are different as some samples contain valid actions but no object references (e.g. ``the man is walking around the kitchen''), and vice versa. 

\begin{table*}[htbp]
\centering
\resizebox{0.95\textwidth}{!} {
\begin{tabular}{ccccc|ccccc}
\hline
\multicolumn{5}{l|}{Video augmentation + original dialogue} & \multicolumn{5}{l}{Video augmentation + no dialogue}   \\
\hline
($\mathcal{I}$, $\mathcal{H}$) & ($\mathcal{I}^-$, $\mathcal{H}$) & ($\mathcal{I}^0$, $\mathcal{H}$) & ($\mathcal{I}^+$, $\mathcal{H}$) & ($\mathcal{I}^-_\mathrm{rand}$, $\mathcal{H}$) & ($\mathcal{I}$, $\mathcal{H}^0$) & ($\mathcal{I}^-$, $\mathcal{H}^0$) & ($\mathcal{I}^0$, $\mathcal{H}^0$) & ($\mathcal{I}^+$, $\mathcal{H}^0$) & ($\mathcal{I}^-_\mathrm{rand}$, $\mathcal{H}^0$) \\
0.779  & 0.760    & 0.733   & 0.782   & 0.773         & 0.724   & 0.708    & 0.693    & 0.722    & 0.710           \\
\hline
\multicolumn{5}{l|}{Dialogue augmentation + original video} & \multicolumn{5}{l}{Dialogue augmentation + no video}   \\
\hline
($\mathcal{I}$, $\mathcal{H}$) & ($\mathcal{I}$, $\mathcal{H}^-$) & (($\mathcal{I}$, $\mathcal{H}^0$) & ($\mathcal{I}$, $\mathcal{H}^+$)  & ($\mathcal{I}$, $\mathcal{H}^-_\mathrm{rand}$) & ($\mathcal{I}^0$, $\mathcal{H}$) & ($\mathcal{I}^0$, $\mathcal{H}^-$) & ($\mathcal{I}^0$, $\mathcal{H}^0$) & ($\mathcal{I}^0$, $\mathcal{H}^+$)  & ($\mathcal{I}^0$, $\mathcal{H}^-_\mathrm{rand}$) \\
0.779  & 0.764   & 0.724   & 0.788   &  0.778             & 0.733   & 0.722    & 0.693    & 0.739    & 0.734         \\
\hline
\end{tabular}
}
\caption{
\textbf{Validation results with augmentation data:}
$\mathcal{I}$: original video input, $\mathcal{I}^{-/+}$: counterfactual/factual video following Section \ref{subsec:cf_augmentation},
$\mathcal{I}^-_\mathrm{rand}$: counterfactual video by masking random temporal steps,
$\mathcal{I}^0$: no video input;
$\mathcal{H}$: original dialogue input, $\mathcal{H}^{-/+}$: counterfactual/factual dialogue following Section \ref{subsec:cf_augmentation}, 
$\mathcal{H}^-_\mathrm{random}$: counterfactual dialouge by masking random tokens, 
$\mathcal{H}^0$: no dialogue input. 
All results are in CIDEr score. 
}
\label{tab:eval_cf}
\end{table*}

\begin{table*}[htbp]
\small
\centering
\begin{tabular}{clllccccccc}
\hline
\# & \multicolumn{1}{c}{\begin{tabular}[c]{@{}c@{}}Contrast\\ pair\end{tabular}} & \multicolumn{1}{c}{\begin{tabular}[c]{@{}c@{}}Contrast \\ loss\end{tabular}} & \multicolumn{1}{c}{\begin{tabular}[c]{@{}c@{}}Hidden \\ states\end{tabular}} & B-1            & B-2            & B-3            & B-4            & M         & R        & C          \\
\hline
A  & -                                                                                 & -                                                                            & -                                                                            & 0.695          & 0.558          & 0.455          & 0.376          & 0.253          & 0.534          & 0.996          \\
\hline
B  & $\mathcal{I}^+$, $\mathcal{I}^-$                                                                            & NCE                                                                          & $D_{act}$                                                                       & \textbf{0.709} & \textbf{0.577} & \textbf{0.476} & \textbf{0.398} & \textbf{0.262} & \textbf{0.549} & \textbf{1.040} \\
\hline
C  & $\mathcal{I}^+$, $\mathcal{I}^-$                                                                            & NCE                                                                          & $D'_{act}$                                                                      & 0.697          & 0.565          & 0.462          & 0.381          & 0.254          & 0.538          & 1.003          \\
D  & $\mathcal{I}^+$, $\mathcal{I}^-$                                                                            & NCE                                                                          & $D_{obj}$                                                                       & 0.701          & 0.565          & 0.462          & 0.383          & 0.256          & 0.541          & 1.011          \\
E  & $\mathcal{I}^+$, $\mathcal{I}^-$                                                                            & NCE                                                                          & $D$                                                                            & 0.699          & 0.566          & 0.465          & 0.386          & 0.253          & 0.539          & 1.008          \\
\hline
F  & $\mathcal{I}^+$, $\mathcal{I}^-_\mathrm{rand}$                                                                        & NCE                                                                          & $D_{act}$                                                                       & 0.693          & 0.563          & 0.464          & 0.388          & 0.254          & 0.538          & 1.010          \\
G  & $\mathcal{I}^+$, $\mathcal{I}^0$                                                                            & NCE                                                                          & $D$                                                                            & 0.700          & 0.566          & 0.463          & 0.383          & 0.256          & 0.538          & 1.019          \\
H  & $\mathcal{I}^+$, $\mathcal{I}^0_\mathrm{rand}$                                                                        & NCE                                                                          & $D$                                                                       & 0.695          & 0.563          & 0.463          & 0.385          & 0.253          & 0.538          & 0.998  \\
\hline
I & $\mathcal{I}^+$, $\mathcal{I}^-$                                                                             & S-NCE                                                                        & $D_{act}$                                                                       & 0.695          & 0.567          & 0.467          & 0.389          & 0.255          & 0.54           & 1.014          \\
J  & $\mathcal{I}^+$, $\mathcal{I}^-$                                                                            & L1-PD                                                                        & $D_{obj}$                                                                       & 0.705          & 0.569          & 0.465          & 0.385          & 0.258          & 0.543          & 1.005          \\
\hline
\end{tabular}
\caption{
\textbf{Contrastive learning with counterfactual videos:}
We experimented with variants of contrastive video pairs, hidden state sampling, and contrast loss. 
Metrics: B-n: BLEU-n, M: METEOR, R: ROUGE-L, C: CIDEr.
}
\label{tab:contrast_video_result}
\end{table*}

\textbf{Evaluating with Counterfactual Data}. 
First, using augmented data, we evaluated models trained only with the original data. 
Motivated by \cite{Kaushik2020Learning, NEURIPS2020_92650b2e, agarwal2020towards}, we designed this set of experiments to gauge the model performance under adversarial (counterfactual) samples and favourable (factual) samples and to observe the effects of our transformation methods. 
Specifically, we trained an MTN model \cite{le-etal-2019-multimodal} on the original training data and evaluate the model on an augmented validation set. 
To fairly compare the results, we create a shared validation set in which each sample is augmented with both video and dialogue factual and counterfactual pairs. 
Essentially, this set is the intersection $\mathrm{Val}^\mathrm{v+d}_\mathrm{aug}=
\mathrm{Val}^\mathrm{video}_\mathrm{aug} \cap \mathrm{Val}^\mathrm{dial}_\mathrm{aug}$. 
Using the CIDEr metric \cite{vedantam2015cider}, We noted the MTN model pretrained on original training data achieves $0.996$ and $1.086$ score in the original test and validation set respectively. 
However, as noted from Table \ref{tab:eval_cf}, the performance drops to $0.779$ when evaluating on the validation set $\mathrm{Val}^\mathrm{v+d}_\mathrm{aug}$ even with the original video-dialogue pair $(\mathcal{I},\mathcal{H})$.
This performance drop indicates that the subset contains more challenging instances that require reasoning in dialogues and videos. 

The performance decreases to $0.760$ when tested with $\mathcal{I}^-$ and increases to $0.782$ when tested with $\mathcal{I}^+$, keeping the $\mathcal{H}$ unchanged.
When tested with videos that are masked at random temporal steps $\mathcal{I}^-_\mathrm{rand}$, the result only reduces to $0.773$, less than $\mathcal{I}^-$.
This illustrates higher counterfactual impacts in $\mathcal{I}^-$ than in $\mathcal{I}^-_\mathrm{rand}$.
We also observed that model performance with counterfactual videos $\mathcal{I}^{-}$ is higher than cases with no video at all, $\mathcal{I}^{0}$.
This observation demonstrates the factorization formulation of our SCM in which $\mathcal{I}^{-}$ is partially counterfactual, containing useful information, i.e. $\mathcal{I}_{obj}$, than $\mathcal{I}^0$, to support response generation. 

When tested with dialogue transformations, we have similar observations with $\mathcal{H}^-$, $\mathcal{H}^+$, $\mathcal{H}^-_\mathrm{rand}$, and $\mathcal{H}^0$.
Specifically, following our SCM structure, we show that $\mathcal{H}^{-}$ is partially counterfactual.
To isolate the impacts of video/dialogue augmentations, we also tested models with tuples that are paired with zero dialogue context/video input ($\mathcal{H}_0$/$\mathcal{I}_0$).
In these isolated experiments, we still observe consistent performance patterns among different variants of augmented video/dialogues, validating our factorization SCM and the effectiveness of augmentation techniques. 

\begin{table*}[htbp]
\small
\centering
\begin{tabular}{clllccccccc}
\hline
\# & \multicolumn{1}{c}{\begin{tabular}[c]{@{}c@{}}Contrast\\ pair\end{tabular}} & \multicolumn{1}{c}{\begin{tabular}[c]{@{}c@{}}Contrast \\ loss\end{tabular}} & \multicolumn{1}{c}{\begin{tabular}[c]{@{}c@{}}Hidden \\ states\end{tabular}} & B-1            & B-2            & B-3            & B-4            & M         & R        & C          \\
\hline
A  & -                                                                                 & -                                                                            & -                                                                            & 0.695          & 0.558          & 0.455          & 0.376          & 0.253          & 0.534          & 0.996          \\
\hline
B  & $\mathcal{H}^+$, $\mathcal{H}^-$                                                                            & NCE                                                                          & $D_{obj}$                                                                       & \textbf{0.705} & \textbf{0.571} & \textbf{0.470} & \textbf{0.393} & \textbf{0.260} & \textbf{0.545} & \textbf{1.029} \\
\hline
C  & $\mathcal{H}^+$, $\mathcal{H}^-$                                                                            & NCE                                                                          & $D'_{obj}$                                                                      &  0.701              & 0.569               & 0.469               & 0.392               & 0.256               &  0.540              &   1.023             \\
D  & $\mathcal{H}^+$, $\mathcal{H}^-$                                                                            & NCE                                                                          & $D_{act}$                                                                       & 0.699          & 0.561          & 0.453          & 0.369          & 0.251          & 0.538          & 0.963          \\
E  & $\mathcal{H}^+$, $\mathcal{H}^-$                                                                            & NCE                                                                          & $D$                                                                            & 0.707          & \textbf{0.571} & 0.466          & 0.385          & 0.258          & 0.542          & 1.020          \\
\hline
F  & $\mathcal{H}^+$, $\mathcal{H}^-_\mathrm{rand}$                                                                        & NCE                                                                          & $D_{obj}$                                                                       & 0.693          & 0.557          & 0.452          & 0.370          & 0.253          & 0.536          & 0.957          \\
G  & $\mathcal{H}^+$, $\mathcal{H}^0$                                                                            & NCE                                                                          & $D$                                                                            & \textbf{0.705} & 0.570          & 0.466          & 0.387          & 0.258          & 0.542          & 1.022          \\
H  & $\mathcal{H}^+$, $\mathcal{H}^0_\mathrm{rand}$                                                                        & NCE                                                                          & $D$                                                                            & 0.696                &       0.563         &  0.462              &   0.383             &  0.254              &   0.536             &   1.005            \\
\hline
I   & $\mathcal{H}^+$, $\mathcal{H}^-$                                                                            & S-NCE                                                                          & $D_{obj}$                                                                       & 0.696          & 0.561          & 0.458          & 0.378          & 0.252          & 0.538          & 0.999          \\
J  & $\mathcal{H}^+$, $\mathcal{H}^-$                                                                            & L1-PD                                                                        & $D_{act}$                                                                       & 0.699          & 0.569          & 0.468          & 0.390          & 0.255          & 0.543          & 1.008          \\ \hline
\end{tabular}
\caption{
\textbf{Contrastive learning with counterfactual dialogues:}
We experiment with variants of contrastive dialogues pairs, hidden state sampling, and loss. 
Metrics: B-n: BLEU-n, M: METEOR, R: ROUGE-L, C: CIDEr.
}
\label{tab:contrast_dial_result}
\end{table*}

\begin{table*}[htbp]
\small
\centering
\begin{tabular}{lclllllll}
\hline
\multicolumn{1}{c}{Model} & \multicolumn{1}{c}{\begin{tabular}[c]{@{}c@{}}Visual \\ Features\end{tabular}} & \multicolumn{1}{c}{B-1}   & \multicolumn{1}{c}{B-2}   & \multicolumn{1}{c}{B-3}            & \multicolumn{1}{c}{B-4}   & \multicolumn{1}{c}{M}              & \multicolumn{1}{c}{R}     & \multicolumn{1}{c}{C}     \\
\hline
Baseline \cite{hori2019avsd}                 & I3D                                                                            & \multicolumn{1}{c}{0.621} & \multicolumn{1}{c}{0.480} & \multicolumn{1}{c}{0.379}          & \multicolumn{1}{c}{0.305} & \multicolumn{1}{c}{0.217}          & \multicolumn{1}{c}{0.481} & \multicolumn{1}{c}{0.733} \\
JMAN \cite{chu2020multi}                     & I3D                                                                            & 0.648                     & 0.499                     & 0.390                              & 0.309                     & 0.240                              & 0.520                     & 0.890                     \\
FA-HRED \cite{nguyen2018film}                   & I3D                                                                            & 0.648                     & 0.505                     & 0.399                              & 0.323                     & 0.231                              & 0.510                     & 0.843                     \\
Student-Teacher \cite{hori2019joint} $\dagger$       &I3D                                                                                & 0.675                     & 0.543                     & 0.446                              & 0.371                     & 0.248                              & 0.527                     & 0.966                     \\
MSTN \cite{lee2020dstc8} $\dagger$                  & I3D                                                                            & -                         & -                         & -                                  & 0.379                     & 0.261                              & 0.548                     & 1.028                     \\
BiST \cite{le-etal-2020-bist}                     & RX                                                                             & \textbf{0.711}            & \textbf{0.578}            & 0.475                              & 0.394                     & 0.261                              & \textbf{0.550}            & 1.050                     \\
RLM-GPT2 \cite{9376902} $\dagger$ $\ddagger$              & I3D                                                                            & \multicolumn{1}{r}{0.694} & \multicolumn{1}{r}{0.570} & \multicolumn{1}{r}{\textbf{0.476}} & \textbf{0.402}            & 0.254                              & 0.544                     & \textbf{1.052}            \\
\hline
MTN \cite{le-etal-2019-multimodal}                & I3D                                                                            & \multicolumn{1}{c}{0.695} & \multicolumn{1}{c}{0.558} & \multicolumn{1}{c}{0.455}          & \multicolumn{1}{c}{0.376} & \multicolumn{1}{c}{0.253}          & \multicolumn{1}{c}{0.534} & \multicolumn{1}{c}{0.996} \\
MTN $+ C^3$ ($\mathcal{I}^{+/-}$)            & I3D                                                                            & \multicolumn{1}{c}{0.709} & \multicolumn{1}{c}{0.577} & \multicolumn{1}{c}{\textbf{0.476}} & \multicolumn{1}{c}{0.398} & \multicolumn{1}{c}{\textbf{0.262}} & \multicolumn{1}{c}{0.549} & \multicolumn{1}{c}{1.040} \\
MTN $+ C^3$ ($\mathcal{H}^{+/-}$)             & I3D                                                                            & \multicolumn{1}{c}{0.705} & \multicolumn{1}{c}{0.571} & \multicolumn{1}{c}{0.470}          & \multicolumn{1}{c}{0.393} & \multicolumn{1}{c}{0.260}          & \multicolumn{1}{c}{0.545} & \multicolumn{1}{c}{1.029}\\
\hline
\end{tabular}
\caption{
\textbf{Overall results}:
$\dagger$ incorporates additional video background audio inputs.
$\ddagger$ indicates finetuning methods on pretrained language models. 
Metrics: B-n: BLEU-n, M: METEOR, R: ROUGE-L, C: CIDEr.
}
\label{tab:overall_result}
\end{table*}

\textbf{Contrastive Learning with Counterfactual Videos}.
In these experiments, we combined the task objective loss with our proposed contrastive learning approach that exploits action-based data contrast between $\mathcal{I}^+$ and $\mathcal{I}^-$.
From Table \ref{tab:contrast_video_result}, we have the following observations: 
\textbf{1)} First, when applying contrastive learning on augmented counterfactual data following our NCE function (\ref{eq:l_act}) (Row B), the model outperforms one which was trained with only original training data (Row A). 
This demonstrates the positive impacts of our $C^3$ learning approach through better generated target responses. 
\textbf{2)} When using the indices of hidden states based on prospective tokens ($D'_{act}$) (Row C), the performance gain decreases. 
This can be explained as hidden states in $D'_{act}$ positions represent contextual information that \emph{potentially}, but not absolutely, generate an action token. 
However, hidden states in  $D_{act}$ positions already assume a hypothetical input action token ($w_{i-1}$ in (\ref{eq:l_act})), and hence, a contrastive learning on these hidden states is more stable. 
\textbf{3)} we observed marginal performance gains when changing hidden state indices to indices of object tokens $D_{obj}$ (Row D) or to hidden states of all tokens $D$ (Row E).
This observation verifies our factorized SCM framework as $\mathcal{I}^-$ and $\mathcal{I}^+$ are formulated to be action-variant specifically. 
Training them based on object variance or generic variance might lead to unstable representation learning and trivial performance gains. 

 \textbf{4)} Consistent with our observations from Table \ref{tab:eval_cf}, contrastive learning applied to counterfactual videos with random masked temporal steps $\mathcal{I}^-_\mathrm{rand}$ (Row F) results in very low performance gain. 
\textbf{5)} When we applied contrastive learning between $\mathcal{I}^+$ and naive counterfactual samples, including 
zero video input $\mathcal{I}^0$ (Row G) or video input sample from other training instance $I^0_\mathrm{rand}$ (Row H), the results only increases marginally compared to results with $\mathcal{I}^-$. 
\textbf{6)}
We experimented with sentence-level contrast (S-NCE) (as in (\ref{eq:sent_nce})) in which all hidden states are considered and collapsed to a single vector, as similarly used by \cite{lee2021contrastive, zhang2020counterfactual}. 
We observed that this loss formulation (Row I), is not effective in our task, illustrating the benefits of using compositional representations of decoding tokens.  
\textbf{7)} Finally, to utilize any object-level invariance between $\mathcal{I}^+$ and $\mathcal{I}^-$, we applied a pairwise L1 distance loss $\mathcal{L}^\mathrm{act}=\sum_i\|sim(z_i,z^+_i) - sim(z_i,z^-_i)\|_1$ to minimizes distances of hidden states of $D_{obj}$ positions (Row J).
However, the performance gain of this loss is not significant, demonstrating representation learning through data variance is a better strategy. 

\textbf{Contrastive Learning with Counterfactual Dialogues}.
From Table \ref{tab:contrast_dial_result}, we observed consistent observations as compared to prior experiments with counterfactual videos. 
Essentially, our results illustrate the impacts of $C^3$ that specifically contrasts object-level information between $\mathcal{H}^-$ and $\mathcal{H}^+$.

\textbf{Overall Results}.
In Table \ref{tab:overall_result}, we reported the results of our models which we trained on an MTN backbone \cite{le-etal-2019-multimodal} incorporated our proposed $C^3$ learning approach with counterfactual videos or dialogues.
Our models achieve very competitive performance against models trained on the same data features e.g. MSTN \cite{lee2020dstc8}, as well as models pretrained with a large language dataset e.g. RLM-GPT2 \cite{9376902}.
We also observed that the performance gain of $C^3$ with $\mathcal{I}^{+/-}$ is higher than that with $\mathcal{H}^{+/-}$. 
As we showed the benefits of augmented counterfactual dialogues and videos, we will leave the study to unify both augmented data types for a hybrid contrastive learning approach for future work. 
In this paper, we showed that either dialogues or videos can be augmented and used to improve contextual representations through contrastive losses based on object-based or action-based variance. 

For example factual/counterfactual videos/dialogues, please refer to the Appendix. 

\section{Discussion and Conclusion}
\label{sec:conclusion}
In this work, we proposed Compositional Counterfactual Contrastive Learning ($C^3$), a contrastive learning framework to address the limitation of data in video-grounded dialogue systems.  
We introduced a factorized object-action structural causal model, described a temporal-based and token-based augmentation process, and formulated contrastive learning losses that exploit object-level and action-level variance between factual and counterfactual training samples. 
In our proposed approach, we train models to minimize the distance between compositional hidden state representations of factual samples and maximize the distance between counterfactual samples.

We noted our proposed $C^3$ still entails some limitations. we describe these limitations and suggest potential ways to overcome them for future extension. 
First, in our approach, we made the assumption of independence between $\mathcal{C}_{obj}$ and $\mathcal{C}_{act}$ to mask tokens/video segments as a way to generate counterfactual data samples. 
However, in many cases, this assumption might be too strong.
Therefore, our approach might disrupt the natural data distribution and create negative noise in model training. 
A more advanced counterfactual data generation should be able to better capture the nature of counterfactual scenarios, avoiding the above assumption and generalizing the model better. 
Secondly, in our approach, we require external text-processing tools to decompose the input components. 
More sophisticated tools could be used to improve data quality of counterfactual/factual examples. 
Finally, after this work was completed, there have been several more advanced approaches following MTN \citep{le-etal-2019-multimodal}.
As our approach is model-agnostic, we encourage readers to review and adapt our work to these more advanced models. 


\section{Broader Impacts}
In this work, we described $C^3$, a novel contrastive learning approach that exploits action-based and object-based variance between counterfactual video/dialogue pairs. 
We demonstrated the benefit of this approach in the video-grounded dialogue domain, which is typically suffered from dataset scarcity. 
We want to emphasize that our method should be used strictly to improve dataset quality and obtain model performance gains. 
For instance, a chatbot that incorporates $C^3$ can generate high-quality responses that better match human questions. 
Our method should not be used for malicious purposes, such as creating chatbots to steal information or make scam calls.  

Considering the widespread application of AI in the real world, the adoption of our method can lead to better dialogue systems that improve the quality of life for many people. 
For instance, a better chatbot embedded in electronic devices will improve both user experience and productivity. 
Conversely, the adoption of dialogue systems might lead to the potential loss of jobs in domains such as customer call centres. 
In high-risk domains such as autonomous vehicles, applications of our method can improve virtual assistant applications in the vehicles.
As the products might directly affect human safety, any applications of $C^3$ should be tested to account for different scenarios, whether the method works as intended or not, and mitigate consequences when the output is incorrect. 
We advise that any plan to apply our method should consider carefully all potential groups of stakeholders as well as the risk profiles of applied domains to maximize the overall positive impacts. 


\bibliography{anthology,custom}
\bibliographystyle{acl_natbib}

\clearpage
\appendix

\begin{figure*}[htbp]
	\centering
	\resizebox{1.0\textwidth}{!} {
	\includegraphics{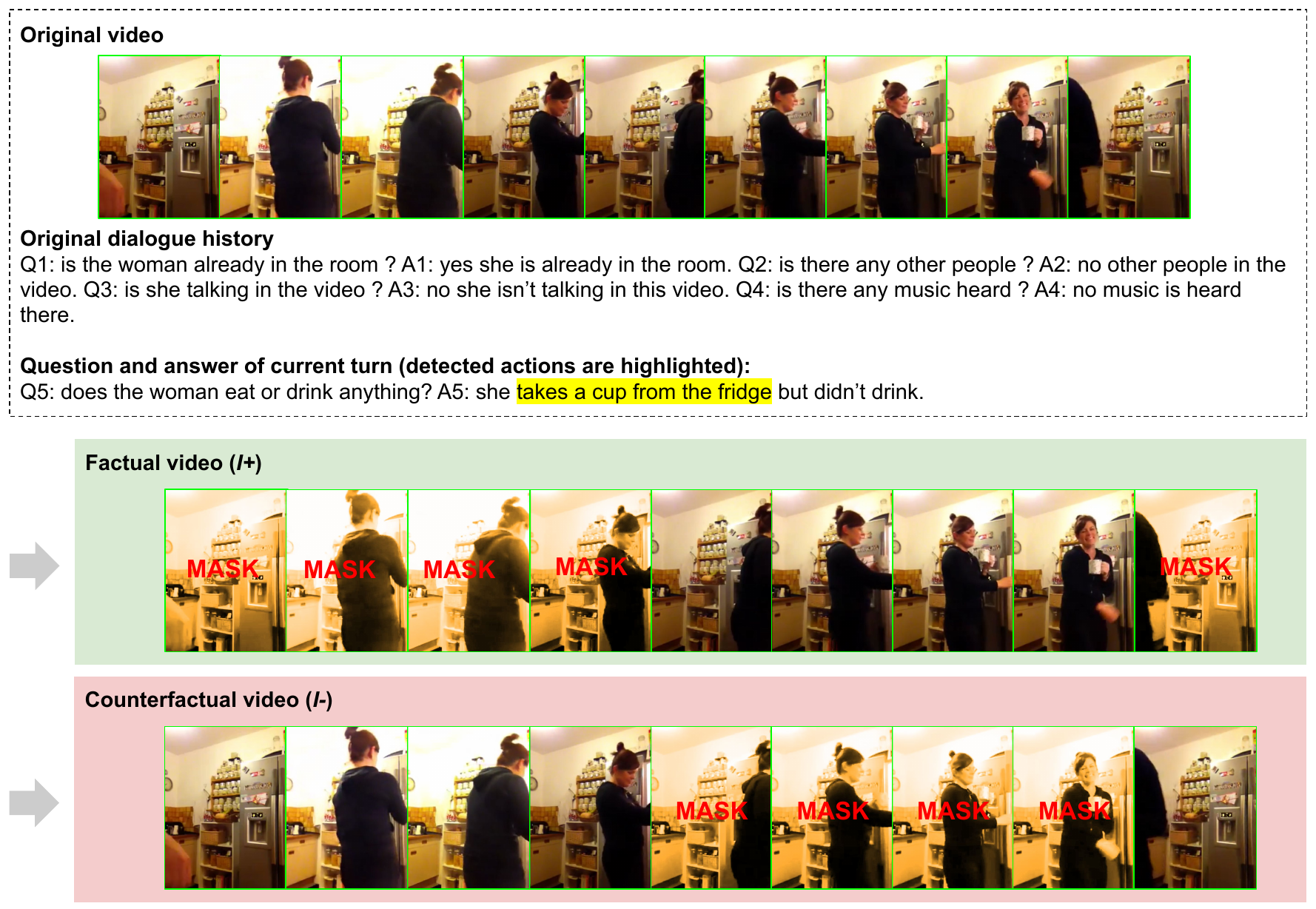}
	}
	\caption{
	Example factual and counterfactual video 
	}
	\label{fig:video1}
\end{figure*}
\begin{figure*}[htbp]
	\centering
	\resizebox{1.0\textwidth}{!} {
	\includegraphics{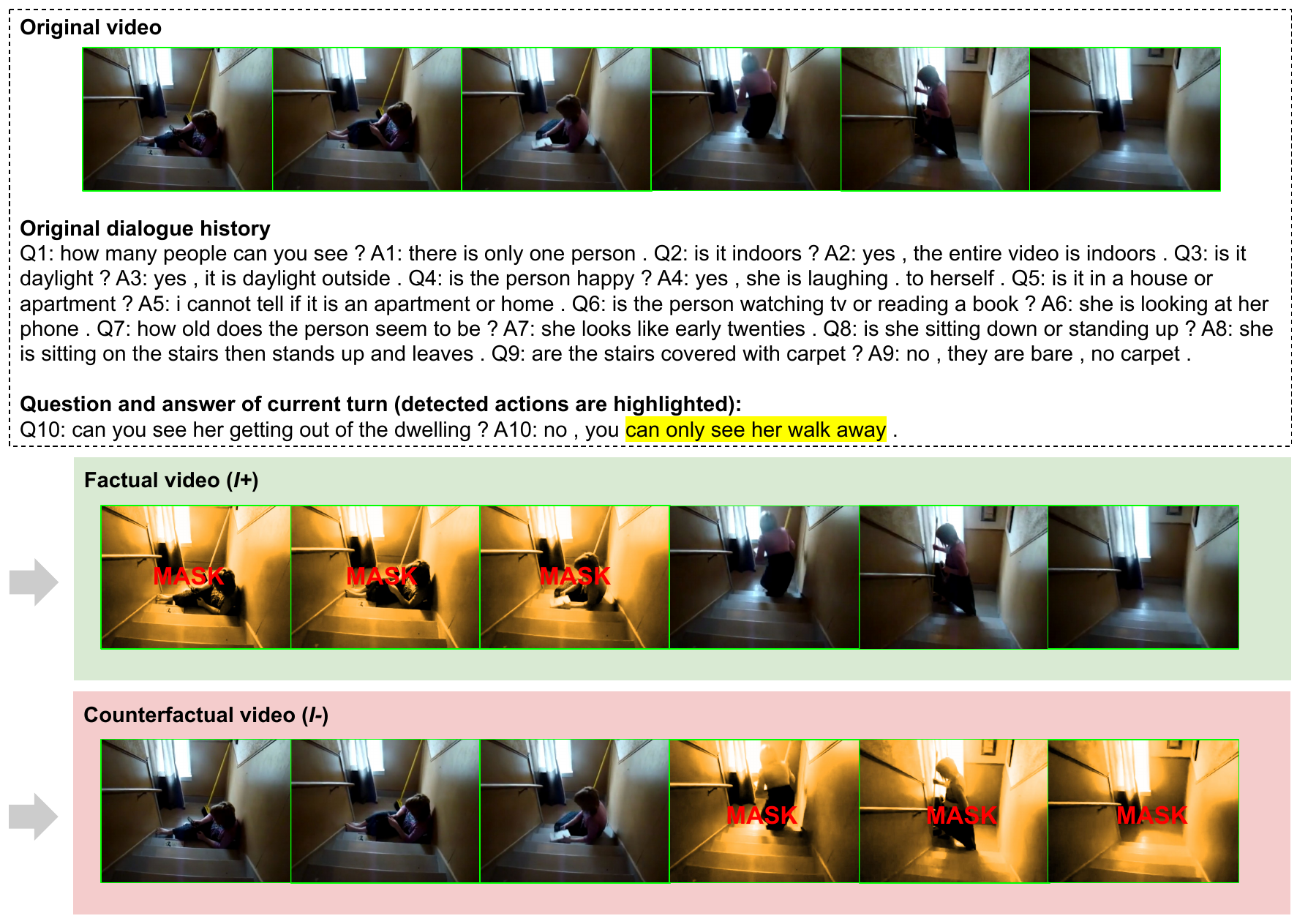}
	}
	\caption{
	Example factual and counterfactual video 
	}
	\label{fig:video2}
\end{figure*}

\begin{figure*}[htbp]
	\centering
	\resizebox{1.0\textwidth}{!} {
	\includegraphics{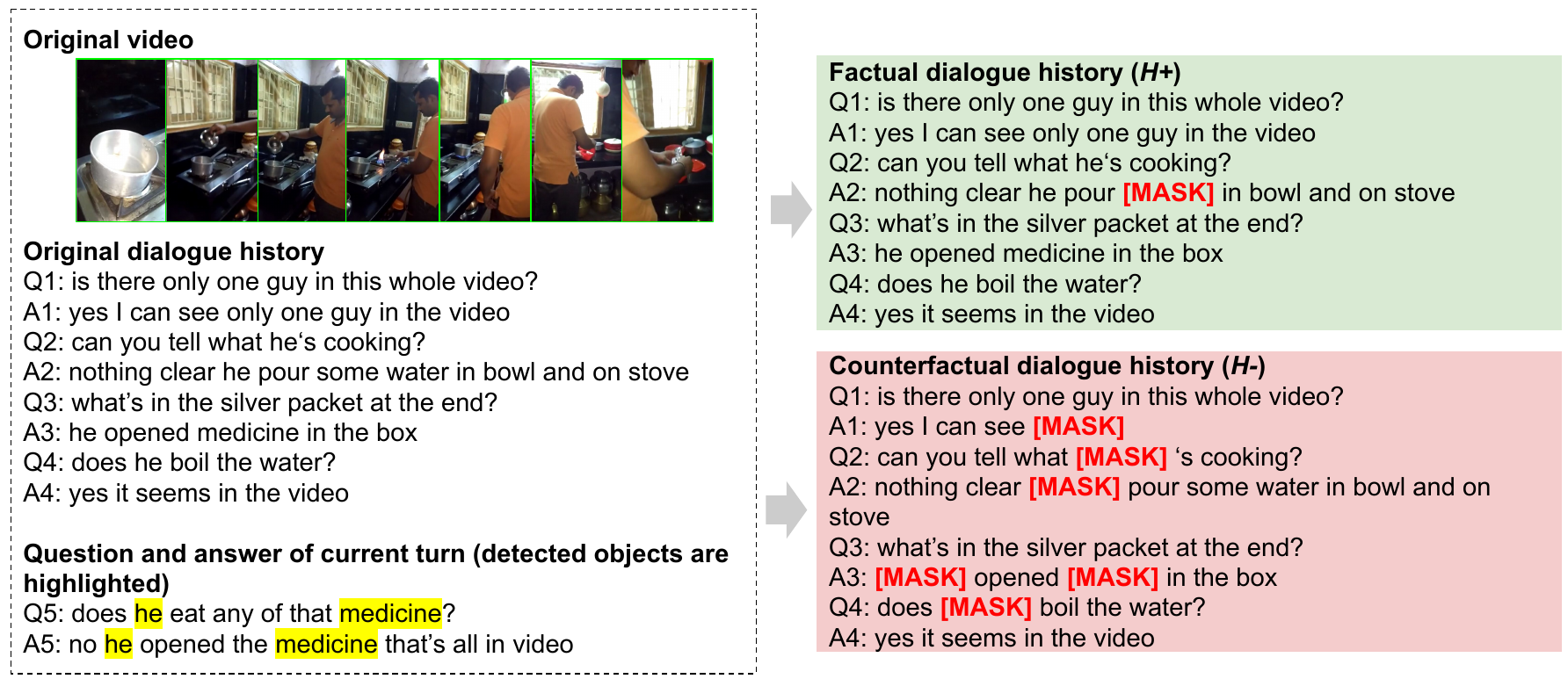}
	}
	\caption{
	Example factual and counterfactual dialogue history 
	}
	\label{fig:dial1}
\end{figure*}
\begin{figure*}[htbp]
	\centering
	\resizebox{1.0\textwidth}{!} {
	\includegraphics{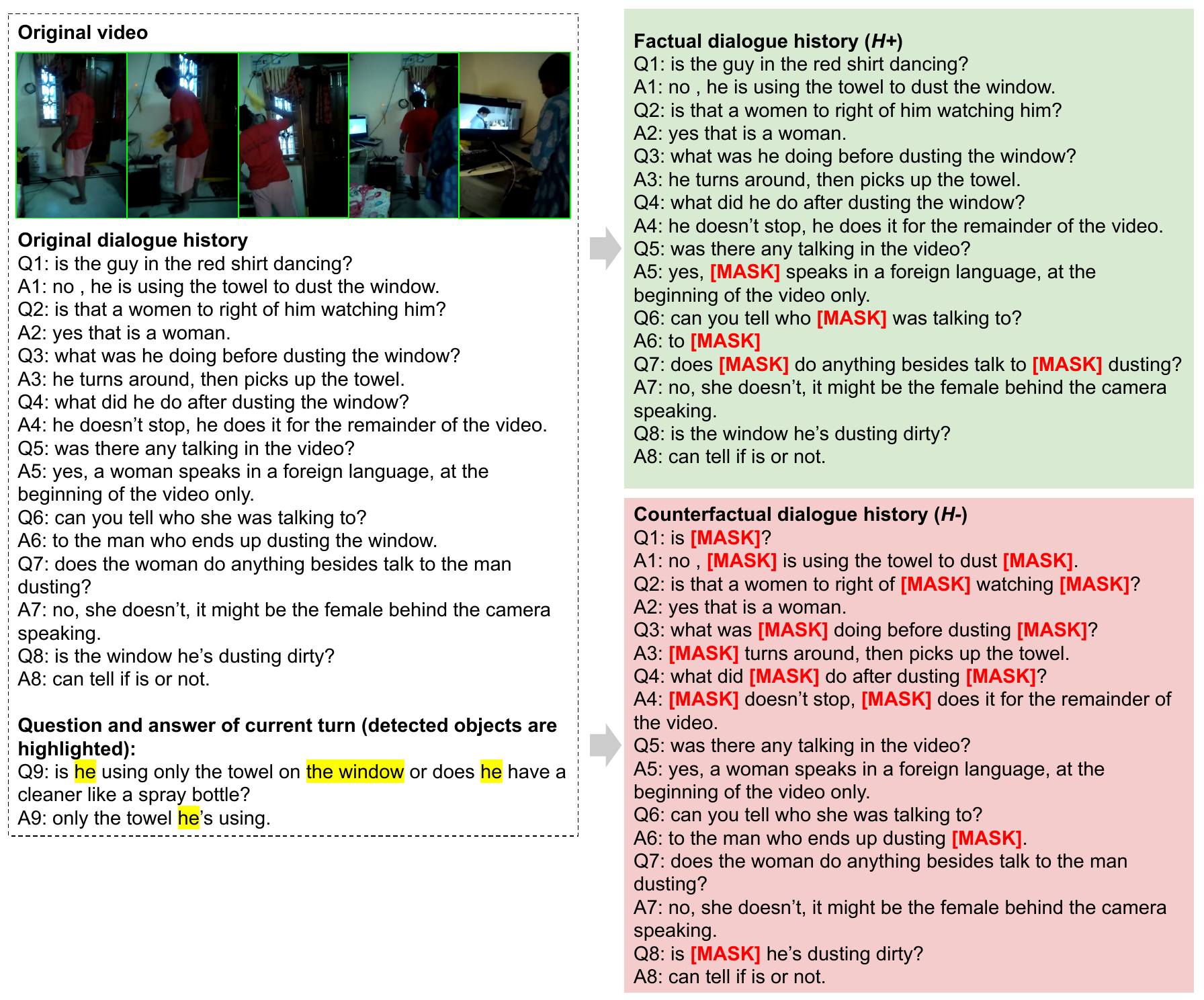}
	}
	\caption{
	Example factual and counterfactual dialogue history 
	}
	\label{fig:dial2}
\end{figure*}

\end{document}